\documentclass[11pt,a4paper]{article}
\usepackage{naaclhlt2018}
\usepackage{times}
\usepackage{latexsym}
\usepackage{graphicx}
\usepackage{subcaption}
\usepackage{xcolor,colortbl}
\usepackage{url}

\aclfinalcopy % Uncomment this line for the final submission
\title{Neural Disease Named Entity Extraction with Character-based BiLSTM+CRF in Japanese Medical Text}

\author{Ken Yano\\
  Nara Institute of Science and Technology \\
  8916-5 Takayama-cho, Ikoma \\
  Nara, Japan \\
  {\tt yano0828@gmail.com} }

\date{}

\begin{document}
\maketitle

\begin{abstract}
We propose an ``end-to-end'' character-based recurrent neural network that extracts disease named entities from a Japanese medical text and simultaneously judges its modality as either positive or negative; i.e., the mentioned disease or symptom is affirmed or negated. The motivation to adopt neural networks is to learn effective lexical and structural representation features for Entity Recognition and also for Positive/Negative classification from an annotated corpora without explicitly providing any rule-based or manual feature sets. We confirmed the superiority of our method over previous char-based CRF or SVM methods in the results.
\end{abstract}

%\begin{abstract}
%In medical text analysis, the disease named entity (DNE) extraction is an important subtask for information extraction and document query. A typical method of named entity (NE) extraction is a cascade of morphological analysis, POS tagging, and chunking. However, for uncommon Japanese named entities such as disease name, the segmentation granularity often contradicts the results of morphological analysis and the building units of NEs, so that correct extraction of such NEs is inherently impossible with this setting. Moreover, in medical texts, the modality of disease name such as affirmed, negated, or doubted, etc. is an important attribute for their statistical analysis and information extraction for post-processing and applications.  We propose a char-based bi-directional LSTMs coupled with CRF for DNE extraction.  From the results, it is confirmed that our methods show superiority over previous char-based CRFs and SVMs. Moreover, unlike previous methods, our method does not need feature engineering efforts such as morphological analysis and it implicitly incorporate a whole sentence to estimate an NE tag, it has the much better ability of correct polarity classification of the extracted DNEs.  The results suggest that the proposed method potentially surpass previous methods in general NE extraction tasks.
%\end{abstract}

\section{Introduction}

In medical text processing, the disease named entity (DNE) extraction is an important task for information extraction and document query. A typical method of a named entity (NE) extraction, in general, is a cascade of morphological analysis, POS tagging, and chunking. However, for uncommon Japanese named entities such as disease name, the segmentation granularity often contradicts the results of morphological analysis and the building units of NEs, so that correct extraction of such NEs is inherently impossible with this setting. Moreover, in medical texts, the modality estimate of DNE such as affirmed, negated, or doubted, etc. is an important subtask for statistical analysis and information retrieval for post-processing medical applications.  In this work, we propose char-based DNE extraction with its modality judgment by using bidirectional LSTM coupled with CRF, which outperforms the `state-of-art' methods.

There have been related works discussing DNE in Japanese medical text \cite{Aramaki2010} \cite{Takeuchi2005} \cite{Miura2013}. Although there are many research articles addressing negation detection for medical texts in English such as NegEx~\cite{Chapman:01a}, NegFinder~\cite{Mutalik:01}, ConText~\cite{Harkema:09}, and \cite{Elkin:05}, they are of little use for analyzing medical text in other languages. Especially for medical text in Japanese, there are few works discussing modality judgment as well as DNE extraction.

\begin{figure}
\begin{center}
\includegraphics[width=0.5\textwidth]{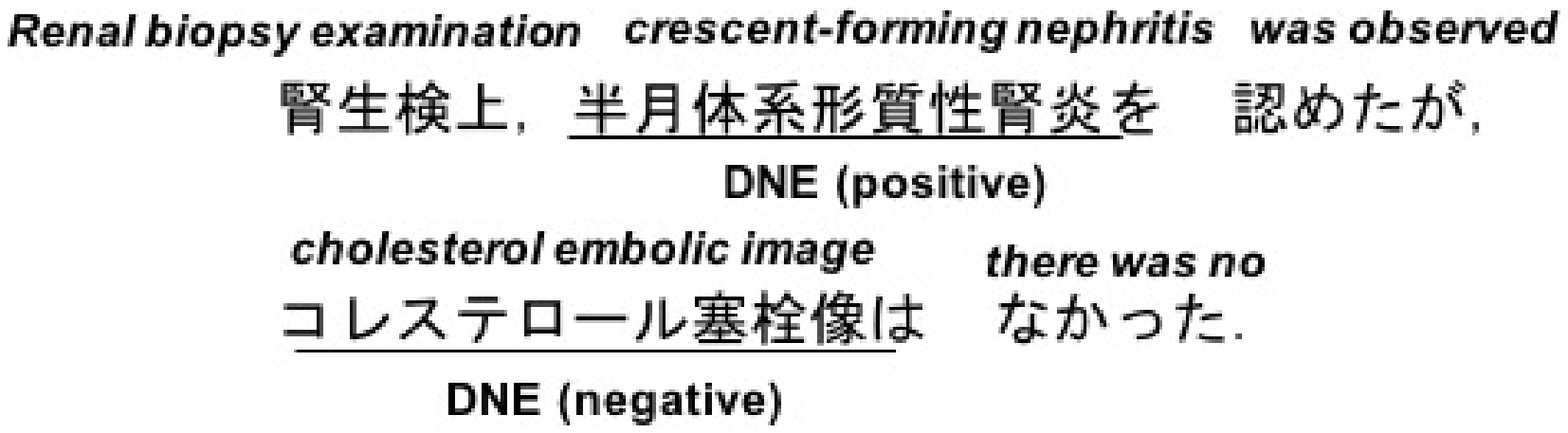}
\end{center}
\caption{DNE extraction with modality judgement}
\label{fig:labeling}
\end{figure}
In this work, we consider the modality as either positive or negative and define the estimation of the modality as P/N classification.
To solve the DNE extraction and it's P/N classification as shown in Fig. \ref{fig:labeling}, we propose an end-to-end character-based recurrent neural network that extracts named entities and simultaneously judges its modality as either positive or negative. The motivation to adopt neural networks is to learn lexical and structural representation features useful for NER extraction and also for P/N classification without explicitly providing any rule-based grammar either in the form of token or syntactic patterns, from a substantially large annotated corpus.

There have been several works discussed the superiority of char-based NE extraction over word-based approaches not only in Japanese but also in other languages, \cite{Chiu:16} \cite{Lample:16} \cite{Kuru:16}.  Especially for Japanese NE extraction, Asahara et al \shortcite{Asahara2003} proposed char-based NE extraction with redundant morphological analysis by using SVM. 
As shown in the results, our proposed method outperforms previous `state-of-art' methods for DNE extraction task.  The advantage of the method is twofold, 1) it does not require morphological and dependency analysis to come up with effective feature sets for NE extraction and 2) it uses the whole sentence to judge the modality of extracted NEs, unlike previous methods that can only consider context information within a fixed window size.

\section{Method}
\begin{figure}
\begin{center}
\includegraphics[width=0.5\textwidth]{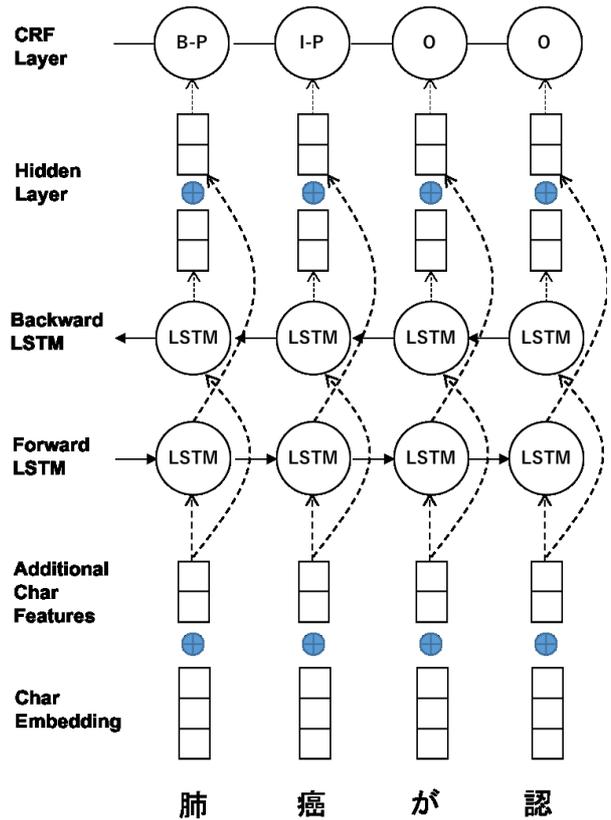}
\end{center}
\caption{Character-based BiLSTM+CRF}
\label{fig:biLSTM-CRF}
\end{figure}

%The proposed method is character-based BiLSTM with CRF. 
%`end-to-end'' character-based recurrent neural network that extracts named entities and simultaneously judges its modality as either positive or negative. 
The Figure \ref{fig:biLSTM-CRF} shows the configuration of the proposed neural network using bi-directional (forward, backward) LSTM coupled with CRF for the output layer. The input at each time step is a character. It is converted to a character vector by the character embedding layer. In addition to the character vector, the input layer can optionally take additional character-based features to leverage the feature representations. We optionally use character type and  ICD (International Statistical Classification of Diseases and Related Health Problems)-10 as for the additional features. These additional features are obtained by preprocessing before training the network. The ICD-10 assignment to names was done by simple string matching using an existing ICD-10 dictionary. Please note that the dictionary contains only Japanese disease name in \textit{standard forms}, so it cannot be used for extracting all possible DNEs in various forms such as abbreviation, etc. which often appear in medical texts.

The concatenated embedding vector of the surface and additional character features are put independently in forwarding and backward LSTM layers. For each time step, the outputs from forward and backward LSTM are concatenated and sent to the hidden layer. The output fully connected with the hidden layer gives the multi-class probability information corresponding to NE tags. Finally, the probability information of NE tags over the entire sequence is put in the CRF layer to estimate the optimal NE tag sequence over the whole text sequence.
The CRF layer estimate NE tags at each char position which jointly maximize the label sequence by the Viterbi algorithm. Table \ref{tbl:parameters} shows parameters of elements constituting each network layer. These values are adjusted to optimal values by a preliminary investigation.

\begin{table}[tb]
\begin{tabular}{l|r|r}
\hline
Layer & Input dim. & Output dim. \\ \hline
Char. Embed & 1516 & 100 \\
Forwad-LSTM & 100 & 100 \\
Backward-LSTM & 100 & 100 \\
Hidden & 200 & 5 \\
CRF & (5,L) & (1,L) \\ \hline
\end{tabular}
\caption{Dimensions of the input and the output of each layer  of BiLSTM+CRF.  The input and the output of CRF are matrixes with specified dimensions.  The $L$ is the length of the sequence in characters. The number of NE tags is five: ``B-P'',``I-P'',``B-N'',``I-N'', and ``O''.}
\label{tbl:parameters}
\end{table}

\subsection{Modality of DNE}
To define the modality of extracted DNE, either positive or negative,  we define a set of rules for annotation. The following are the sample rules which determine the DNE as negative. If none of the following predicates can be acknowledged, the DNE is determined as positive.  

The DNE is 1) stated/annotated as negative or 2) not developed yet or 3) mentioned in the family history or 4) related to anamnesis or 5) acknowledged as healed.

%\begin{itemize}
%\item The DNE is affirmed as negative.
%\item The DNE is not developed yet.
%\item The DNE is mentioned in the family history.
%\item The DNE is related to anamnesis.
%\item The DNE is acknowledged as healed.
%\end{itemize}

In the training corpora, positive DNEs are labeled as P-tag and negative DNEs are labeled as N-tag.   Since we adopt IOB2 sequence label, the P-tag is labeled by ``B-P'' and ``I-P'', likewise the N-tag is labeled by ``B-N'', ``I-N''. The characters that do not belong to DNE are indicated by the ``O'' label.

For each rule of negative DNE described above as well as for positive DNE, there are many various forms of mentions in medical texts, hence it is difficult to describe them by rule-based approaches.  The proposed method tries to solve it by implicitly learning mention patterns of positive or negative DNEs from the annotated corpus.

In our preliminary experiments, we confirmed that consistent IOB2 labeling results are not always obtained when a softmax layer is used instead of CRF after the hidden layer as shown in Fig. \ref{fig:biLSTM-CRF}. For example, after the "B-P" label, either "I-P" or "O" should appear, but "I-N" sometimes appear when using the softmax layer. The CRF enables learning of the transition probabilities between NE tags, thereby eliminating the impossible tag sequences. 

\section{Experiments}
This study used a set of 500 annotated medical case reports (5,158 sentences in total) which is compatible with NTCIR shared task data~\cite{Morita13} as the corpus. In these data,  P-tag and N-tag were manually annotated by an annotator and later it is checked by one with medical knowledge. 
The agreement-score of P-tag and N-tag are 81.8 and 67.0 respectively in $F_{\beta=1}$ scale, which suggests the negative decision is vulnerable due to the ambiguity of mention patterns.

\subsection{Evaluation}
The evaluation was performed by ten-fold cross validation by replacing the test set from 10 divided corpora. The final performance evaluation was calculated by averaging the 10 results from each fold.

The evaluation method is the same as that of the CoNLL-2000 shared task. The Perl script used for evaluation is obtainable from the CoNLL-2000 Website \footnote{https://www.clips.uantwerpen.be/conll2000/}. The performance was measured using precision, recall, and $F_{\beta=1}$ scores.

\subsection{Char-based Features}
\begin{figure}[tb]
\begin{center}
\includegraphics[width=0.5\textwidth]{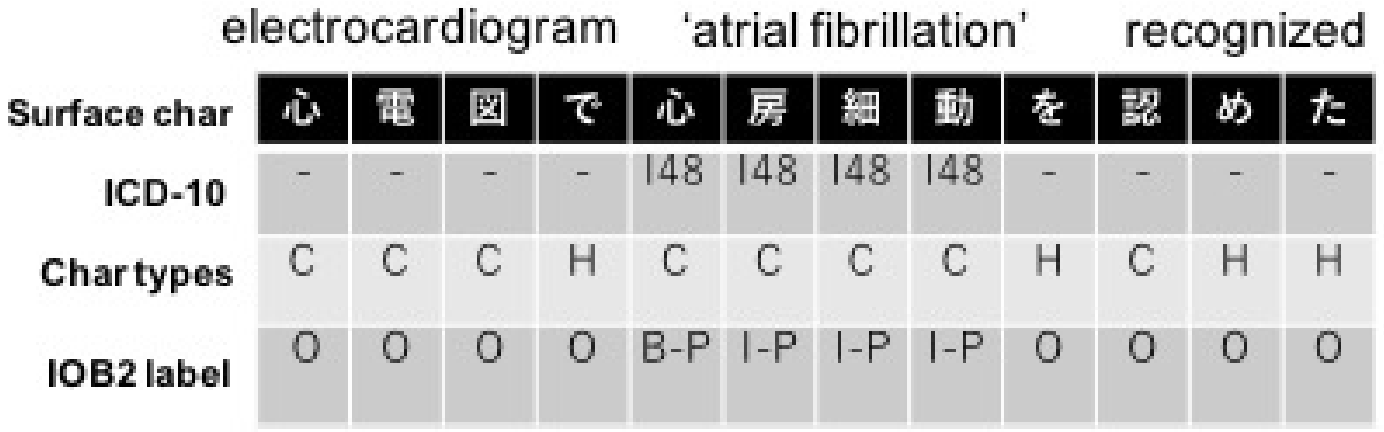}
\end{center}
\caption{Char-based input text and its attributes. From top to bottom: input text, ICD-10 code, Japanese character types (C (kanji), H (hiragana), K (katakana), A(alpha numeric)), and IOB2 sequence label. I48 is ICD-10 code for atrial fibrillation.}
\label{fig:input_feature}
\end{figure}

Figure \ref{fig:input_feature} presents character features used for the method. 
For experiments, we used four different feature sets: 1) use only surface characters (BiLSTM), 2) use char type as additional feature (BiLSTM\_ct) , 3) use ICD-10 code as additional feature  (BiLSTM\_icd), and 4) use both ICD-10 code and char type as additional features  (BiLSTM\_ct\_icd).

Assigning char type and ICD-10 code was done in the preprocessing process. TFfable \ref{tbl:dictionary} presents the input character information, the size of each dictionary, and the dimension of the transformed embedding vectors.

\begin{table}[tb]
\begin{tabular}{l|r|r}
\hline
Feature & Dic. size & Dim. of Embed Vec. \\ \hline
Char. & 1516 & 100 \\
ICD-10 & 15040 & 100 \\
Char. type & 4 & 10 \\ \hline
\end{tabular}
\caption{Character features with its the entry size and the corresponding embedding vector size.}
\label{tbl:dictionary}
\end{table}

\subsection{Training the Network}
We used chainer\footnote{https://chainer.org/} to implement the network and set the training parameters as follows;
minibatch size: 10 (randomize the sequence of input text in each epoch), dropout: 0.5 (linearly downscale to 0 as epoch proceeds)
optimizer: Adam \cite{Kingma2015}.

The objective loss function to minimize is defined as the negative log-likelihood of the CRF layer which computes the joint probability of  NE tag sequence over a sentence.

\section{Results}

\begin{table*}[tb]
\begin{tabular}{c||c|c|c|c||c}
Position & Char. & Char. Type &  POS(Best) & POS(2nd) & NE tag \\ \hline
i-2 & $\surd$ & $\surd$ & $\surd$ & $\surd$ & * \\
i-1 & $\surd$(2) & $\surd$ & $\surd$ & $\surd$ & * \\
i    & $\surd$(1),(2) & $\surd$ & $\surd$ & $\surd$ & ?  \\
i+1 & $\surd$ & $\surd$ & $\surd$ & $\surd$ &  \\
i+2 & $\surd$ & $\surd$ & $\surd$ & $\surd$ &  \\
\end{tabular}
\caption{Char-based features used for CRF and SVM methods.  To estimated NE tag at position $i$ denoted by `?',  SVM and CRF use all the features denoted by `$\surd$'.  In addition, SVM uses estimated NE tags denoted by `*'. CRF\_unigram and CRF\_bigram only use Char. features denoted by (1) and (2) as the baselines.}
\label{tbl:char-features}
\end{table*}

\begin{table*}[tbh]
\begin{tabular}{l|r|r|r|r|r|r|r|r|r}
%& \multicolumn{3}{c}{P-tag} & \multicolumn{3}{c}{N-tag} & \multicolumn{3}{c}{ER} \\ \hline
%Word-based   & Precision& Recall & $F_{\beta=1}$ & Precision& Recall & $F_{\beta=1}$  & Precision& Recall & $F_{\beta=1}$  \\ \hline
%Unigram & 75.9 & 74.4 & 75.1 & 37.2 & 14.1 & 20.3 & 85.4 & 76.6 & 80.7 \\
%Bigram & 72.8 & 45.1 & 55.6 & 35.8 & 9.7 & 15.1 & 80.8 & 46.0 & 58.6 \\
%CRF  & 84.9 & 81.6 & 83.2 & 60.9 & 39.2 & 47.6 & 91.4 & 83.6 & 87.3 \\ 
%SVM & 85.5 & 81.0 & 83.2 & 63.3 & 37.3 & 46.6 & 91.4 & 82.0 & 86.4 \\
%BiLSTM	 & 85.2 & 59.2 & 69.8  & 74.6 & 41.9 & 52.9 & 89.0 & 60.1 & 71.7 \\ \hline
%
& \multicolumn{3}{c}{DNE(P-tag)} & \multicolumn{3}{c}{DNE(N-tag)} & \multicolumn{3}{c}{DNE-E} \\ \hline
Method   & Precision& Recall & $F_{\beta=1}$ & Precision& Recall & $F_{\beta=1}$  & Precision& Recall & $F_{\beta=1}$  \\ \hline \hline
CRF\_unigram & 72.5 & 68.4 & 70.4 & 37.9 & 11.5 & 17.5 & 81.3 & 69.4 & 74.9 \\
CRF\_bigram  & 72.9 & 57.2 & 64.1 & 30.9 & 10.5 & 15.6 & 80.5 & 58.1 & 67.5 \\
CRF & 83.3 & 81.6 & 82.4 & 58.6 & 41.2 & 48.3 & 89.9 & 84.6 & 87.2 \\
SVM  & 83.6 & 82.6 & 83.1 & 58.3 & 44.5 & 50.1 & 89.7 & 85.8 & 87.7 \\
BiLSTM & 86.6 & 86.5 & 86.5 & 73.1 & 58.0 & 64.1 & 89.7 & 87.1 & 88.4 \\
BiLSTM\_ct  & 86.9 & 87.7 & 87.3  & 77.8 & 54.9 & 64.2 & 90.4 & 87.2 & 88.7 \\
BiLSTM\_icd   & 88.7 & 87.9 & 88.3 & 73.5 & \textbf{58.2} & \textbf{64.7} & 91.8 & 88.4 & 90.1 \\
BiLSTM\_ct\_icd & \textbf{88.8} & \textbf{88.4} & \textbf{88.6} &  \textbf{76.3} & 56.8 & 64.5 & \textbf{92.3} & \textbf{88.7} & \textbf{90.5} \\ \hline
\end{tabular}
\caption{Comparative results of DNE(P-tag) and DNE(N-tag) extraction. CRF\_unigram and CRF\_bigram are the baseline methods. The SVM is based on Asahara et al. \shortcite{Asahara2003} which uses the second degree of polynomial kernel. The BiLSTMs are the proposed methods with optionally additional character features; character type or/and  ICD-10.}
\label{tbl:performance}
\end{table*}

Table \ref{tbl:performance} presents the comparative results of DNE extraction among different char-based methods.
CRF\cite{Lafferty:01} are implemented by using CRF++ \footnote{https://taku910.github.io/crfpp/}. For SVM we adopt yamcha \cite{Kudo:01} and use the second degree of the polynomial for the kernel and backward for the parsing direction since it shows better performance than forwarding direction.
Table \ref{tbl:char-features} shows the char-based feature used for CRF and SVM as mentioned in the work by Asahara \& Matsumoto 2003.   As described in \cite{Asahara2003}, we adopt SE tag model to encode character position along with POS tag to represent character n-best POS features.
To obtain DNE-E (DNE Extraction) performance, we ignored the modality of DNE, either positive or negative, by replacing those extracted DNE tags with a common tag. 
The CRF\_unigram and CRF\_bigram only use unigram and bigram Char. features to estimate NE tags in order to show the performance of baselines. For all the CRF methods, we assume first-order Markov chain model for NE tags. 
BiLSTM using only Char. feature outperforms the CRFs and SVM in all metrics.  Especially for N-tag, it improves more than 14.0 point in $F_{\beta=1}$ scale with regard to the previous methods.  Since the method uses the whole sentence as a relevant context in order to judge the modality of DNE, it outperforms the other that only consider a context of fixed size.  
In Welch's t-test, we confirmed significant higher performance in $F_{\beta=1}$ of BiLSTM against CRF with p=0.0004 for DNE(P-tag), p=0.0002 for DNE(N-tag), and against SVM with p=0.0003 for DNE(P-tag), p=0.0007 for DNE(N-tag).  However, as for DNE-E we only confirmed the significantly higher performance of BiLSTM over CRF and SVM when additional features ICD-10 is used.
When using additional char-features, char-type or ICD-10, a slight performance improvement was observed over BiLSTM, but the difference was not as great as that of between the proposed method and previous methods.
It is also noted that the performance of the N-tag is much lower than that of the P-tag. There are two assumptions for this; one is that negative decision can become a rather difficult task even for humans as indicated by annotation agreement scores and the other is that the frequency of N-tag (14\%) is much lower than P-tag (86\%), which hinders the performance of N-tag with limited training corpus.

\section{Conclusion}
In this work, we propose an ``end-to-end'' character-based recurrent neural network that extracts disease named entities from a medical text and simultaneously judges its modality as either positive or negative. 
The proposed method has two important benefits: 1) simplified processing due to the lack of requirement for morphological analysis and dependency parsing, and 2) automatic learning of relevant features effective for DNE extraction and its modality judgment as either positive or negative.
%We also speculate that the method is effective not only for DNE but also for NE extraction in general.

\section*{ Acknowledgments}
Part of this research was supported by Grants-in-Aid from the Japan Society for the Promotion of Science Grant Nos. JP 16H 06395 and 16 H 06399, and by a Grant-in-Aid for Scientific Research Grant No. 28030301.

%\bibliography{ijcnlp2017_yano}

\begin{thebibliography}{}
\expandafter\ifx\csname natexlab\endcsname\relax\def\natexlab#1{#1}\fi

\bibitem[{Aramaki et~al.(2010)Aramaki, Miura, Tonoike, Ohkuma, Masuichi, Waki,
  and Ohe}]{Aramaki2010}
Eiji Aramaki, Yasuhide Miura, Masatsugu Tonoike, Tomoko Ohkuma, Hiroshi
  Masuichi, Kayo Waki, and Kazuhiko Ohe. 2010.
\newblock {Extraction of adverse drug effects from clinical records}.
\newblock {\em Studies in Health Technology and Informatics\/} 160(PART
  1):739--743.

\bibitem[{Asahara and Matsumoto(2003)}]{Asahara2003}
Masayuki Asahara and Yuji Matsumoto. 2003.
\newblock {Japanese Named Entity extraction with redundant morphological
  analysis}.
\newblock In {\em Proc. of NAACL\/}. pages 8--15.

\bibitem[{Chapman et~al.(2001)Chapman, Bridewell, Hanbury, Cooper, and
  Buchanan}]{Chapman:01a}
Wendy~W. Chapman, Will Bridewell, Paul Hanbury, Gregory~F. Cooper, and Bruce~G.
  Buchanan. 2001.
\newblock {A Simple Algorithm for Identifying Negated Findings and Diseases in
  Discharge Summaries}.
\newblock {\em Journal of Biomedical Informatics\/} 34(5):301--310.

\bibitem[{Chiu and Nichols(2016)}]{Chiu:16}
Jason~P.C. Chiu and Eric Nichols. 2016.
\newblock {{Named Entity Recognition with Bidirectional LSTM-CNNs}}.
\newblock {\em Trans. of the ACL\/} 4(5):357--370.

\bibitem[{Elkin et~al.(2005)Elkin, Brown, Bauer, Husser, Carruth, Bergstrom,
  and Wahner-Roedler}]{Elkin:05}
Peter~L. Elkin, Steven~H. Brown, Brent~A. Bauer, Casey~S. Husser, William
  Carruth, Larry~R. Bergstrom, and Dietlind~L. Wahner-Roedler. 2005.
\newblock {A controlled trial of automated classification of negation from
  clinical notes}.
\newblock {\em BMC Medical Informatics and Decision Making\/} 8(6).

\bibitem[{Harkema et~al.(2009)Harkema, Dowling, Thornblade, and
  Chapman}]{Harkema:09}
Henk Harkema, John~N. Dowling, Tyler Thornblade, and Wendy~W. Chapman. 2009.
\newblock {Context: An Algorithm for Determining Negation, Experiencer, and
  Temporal Status from Clinical Reports}.
\newblock {\em Journal of Biomedical Informatics\/} 42(5):839--851.

\bibitem[{Kingma and Ba(2015)}]{Kingma2015}
Diederik~P. Kingma and Jimmy Ba. 2015.
\newblock {Adam: A Method for Stochastic Optimization}.
\newblock In {\em Proc. of 3rd International Conference on Learning
  Representations\/}.

\bibitem[{Kudo and Matsumoto(2001)}]{Kudo:01}
Taku Kudo and Yuji Matsumoto. 2001.
\newblock {Chunking with Support Vector Machines}.
\newblock In {\em Proc. of NAACL\/}.

\bibitem[{Kuru et~al.(2016)Kuru, Can, and Yuret}]{Kuru:16}
Onur Kuru, Ozan~A. Can, and Deniz Yuret. 2016.
\newblock {CharNER: Character-Level Named Entity Recognition}.
\newblock In {\em Proc. of COLING\/}. pages 911--921.

\bibitem[{Lafferty et~al.(2001)Lafferty, McCallum, and Pereira}]{Lafferty:01}
John Lafferty, Andrew McCallum, and Fernando Pereira. 2001.
\newblock {Conditional random fields: Probabilistic models for segmenting and
  labeling sequence data}.
\newblock In {\em Proc. of the International Conference on Machine Learning\/}.
  volume~18, pages 282--289.

\bibitem[{Lample et~al.(2016)Lample, Ballesteros, Subramanian, Kawakami, and
  Dyer}]{Lample:16}
Guillaume Lample, Miguel Ballesteros, Sandeep Subramanian, Kazuya Kawakami, and
  Chris Dyer. 2016.
\newblock {Neural Architectures for Named Entity Recognition}.
\newblock In {\em Proc. of NAACL-HLT\/}. pages 260--270.

\bibitem[{Miura(2013)}]{Miura2013}
Yasuhide Miura. 2013.
\newblock {Incorporating Knowledge Resources to Enhance Medical Information
  Extraction}.
\newblock {\em International Joint Conference on Natural Language Processing
  Workshop on Natural Language Processing for Medical and Healthcare Fields\/}
  (October):1--6.

\bibitem[{Morita et~al.(2013)Morita, Kano, Ohkuma, Miyabe, and
  Aramaki}]{Morita13}
Mizuki Morita, Yoshinobu Kano, Tomoko Ohkuma, Mai Miyabe, and Eiji Aramaki.
  2013.
\newblock {Overview of the NTCIR-10 MEDNLP task}.
\newblock In {\em In Proceedings of NTCIR-10\/}.

\bibitem[{Mutalik et~al.(2001)Mutalik, Deshpande, and Nadkarni}]{Mutalik:01}
Pradeep~G. Mutalik, Aniruddha Deshpande, and Prakash~M. Nadkarni. 2001.
\newblock {Use of General-purpose Negation Detection to Augment Concept
  Indexing of Medical Documents}.
\newblock {\em J. Am. Med. Inform. Assoc.\/} 5(13):598--609.

\bibitem[{Takeuchi and Collier(2005)}]{Takeuchi2005}
Koichi Takeuchi and Nigel Collier. 2005.
\newblock {Bio-medical entity extraction using support vector machines}.
\newblock {\em Artificial Intelligence in Medicine\/} 33(2):125--137.

\end{thebibliography}

\bibliographystyle{acl_natbib}

\end{document}